%% file: main.tex
\documentclass{article}
\usepackage{algorithm}
\usepackage{algorithmic}
\usepackage{amsmath}
\usepackage{caption}
\usepackage{graphicx}
\usepackage{microtype}
\usepackage{pgfplots}
\usepackage{spconf}
\usepackage{subcaption}
\usepackage{tikz}
\usepackage{hyperref}

\input{math_commands}

\title{Parallel Composition of Weighted Finite-State Transducers}

\twoauthors
  {Shubho Sengupta, Vineel Pratap}{Facebook AI Reseearch}
  {Awni Hannun}{Zoom AI}

\begin{document}
\maketitle

\begin{abstract}
    Finite-state transducers (FSTs) are frequently used in speech recognition.
    Transducer composition is an essential operation for combining different
    sources of information at different granularities. However, composition is
    also one of the more computationally expensive operations. Due to the
    heterogeneous structure of FSTs, parallel algorithms for composition are
    suboptimal in efficiency, generality, or both. We propose an algorithm for
    parallel composition and implement it on graphics processing units. We
    benchmark our parallel algorithm on the composition of random graphs and
    the composition of graphs commonly used in speech recognition. The
    parallel composition scales better with the size of the input graphs and
    for large graphs can be as much as 10 to 30 times faster than a
    sequential CPU algorithm.
\end{abstract}


\input{introduction}
\input{related}
\input{parallel_composition}

\input{experiments}

\input{conclusion}

\pagebreak

\bibliographystyle{IEEEbib}
\bibliography{refs}

\end{document}

%% file: math_commands.tex
\newcommand{\vx}{{\bf x}}
\newcommand{\vy}{{\bf y}}
\newcommand{\vz}{{\bf z}}

\newcommand{\gA}{\mathcal{A}}
\newcommand{\gB}{\mathcal{B}}
\newcommand{\gC}{\mathcal{C}}



%% file: introduction.tex
\section{Introduction}
\label{sec:introduction}

Finite-state transducers (FSTs) are widely used in speech recognition, natural
language processing, optical character recognition, and other
applications~\cite{breuel2008ocropus, knight2009applications, mohri1997finite,
mohri2002weighted}. A primary function of FSTs is to combine information from
different sources at different granularities. In speech recognition, for
example, FSTs are used to combine word-level language models, phoneme-level
lexicons, and acoustic models operating on sub-phonetic states. This
combination is done by composing FSTs representing each source.

Composition is arguably the most important operation used with FSTs. It is also
one of the most expensive. If the input FSTs have $V_1$ and $V_2$
nodes, and maximum out-degrees (outgoing arcs per node) of $D_1$ and $D_2$, then
the composition scales as $O(V_1V_2D_1D_2)$. Furthermore, the heterogeneity of
FSTs makes it difficult to implement a parallel composition.

In speech recognition, FSTs are primarily used for inference in the decoder. In
these cases, running time and space usage can be greatly improved by either
pre-computing the FST compositions or with on-the-fly or lazy
implementations~\cite{hori2005generalized, mohri2008speech}. However, with the
advent of automatic differentiation with automata and their use at training
time~\cite{hannun2020dwfst, k2}, constructing the full composed graph is
important. This implies an unmet need for an efficient, eager implementation of
FST composition.

We propose a SIMD-style parallel algorithm for FST composition on GPUs. Our
algorithm supports arbitrary graph structures and allows for $\epsilon$
transitions. We validate our algorithm on two benchmarks: 1) the composition of
two random graphs and 2) the composition of a lexicon graph and an emissions
graph. The latter approximates an operation commonly used in speech recognition
decoding. For large graphs, the GPU composition is between 10 and
30 times faster than a highly optimized sequential implementation.

%% file: related.tex
\section{Related Work}
\label{sec:related}

This work builds on a large body of research in algorithms for efficient FST
composition. In some cases, heuristics such as lazy, on-the-fly, or dynamic
composition combine graph search with composition precluding the need to
construct the full composed graph~\cite{cheng2007generalized,
hori2005generalized, ljolje1999efficient, mohri2008speech}. However, when using
FSTs to train models the complete graph is usually required. Parallel
implementations which construct the full composed graph have been developed
for multiple CPUs~\cite{jung2017parallel, jurish2013multi,
mytkowicz2014data}. However, due to differing constraints and opportunities for
parallelism, CPU and GPU devices require different implementations.

This work also builds on research in parallel implementations of core graph
operations on GPUs. Examples include breadth-first
search~\cite{hong2011accelerating, merrill2012scalable}, single-source shortest
path algorithms~\cite{davidson2014work}, and all-pairs shortest path
algorithms~\cite{harish2007accelerating}. However, little prior work exists
exploring FST composition on GPUs. Argueta and
Chiang~\cite{argueta2018composing} developed a GPU composition without support
for $\epsilon$ transitions and which outperformed sequential baselines on a toy
machine-translation task. In contrast, our GPU composition allows for
$\epsilon$ transitions and improves over sequential baselines in more practical
settings.

%% file: parallel_composition.tex
\section{Parallel Composition}
\label{sec:parallel_composition}

A weighted finite-state transducer is a graph which maps input sequences $\vx$
to output sequences $\vy$ with a corresponding score. We denote by $\gA(\vx,
\vy)$ the score with which the weighted FST $\gA$ transduces $\vx$ to $\vy$. We
assume weights are in the log semiring, hence, the score of a path in a
transducer is the sum of the weights along the individual edges. However, the
following composition algorithms generalize easily to other semirings.

\subsection{Composition}

Assume $\gA$ transduces $\vx$ to $\vy$ with score $\gA(\vx, \vy)$. Assume also
that $\gB$ transduces $\vy$ to $\vz$ with score $\gB(\vy, \vz)$. The
composition $\gC$ of $\gA$ and $\gB$ transduces $\vx$ to $\vz$ with the
score given by:
\begin{equation}
    \gC(\vx, \vz) = \sum_\vy \gA(\vx, \vy) + \gB(\vy, \vz).
\end{equation}
Our sequential composition is outlined in
algorithm~\ref{alg:sequential_compose}. The algorithm yields trim graphs.  This
means every state is both accessible (\emph{i.e.} reachable from a start state)
and co-accessible (\emph{i.e.} can reach an accept state). In order to yield
trim graphs, the algorithm proceeds in two stages. The first step is to compute
the set of co-accessible states in the composed graph. This is done by calling
a subroutine in line~\ref{alg:line:coaccessible}. The co-accessible subroutine
is nearly identical to the main body of algorithm~\ref{alg:sequential_compose},
but proceeds backwards from the accept states. The second step is to compute
the set of accessible states while only retaining those which are also
co-accessible.

\begin{algorithm}[t]
\caption{Sequential Composition}
\label{alg:sequential_compose}
\begin{algorithmic}[1]
\STATE \textbf{Input}: Transducers $\gA$ and $\gB$
\STATE Initialize the queue $Q$ and the composed graph $\gC$.
\STATE Compute $R$, the set of co-accessible states in $\gC$.  \label{alg:line:coaccessible}

\FOR {$s_a$ and $s_b$ in all start state pairs of $\gA$ and $\gB$}
    \IF {$(s_a, s_b)$ is in $R$}
        \STATE Add $(s_a, s_b)$ to $Q$ and as a start state in $\gC$.
        \IF {$s_a$ and $s_b$ are accept states}
            \STATE Make $(s_a, s_b)$ an accept state in $\gC$.
        \ENDIF
    \ENDIF
\ENDFOR

\WHILE {$Q$ is not empty} \label{alg:line:state_loop}
    \STATE Remove the next state pair $(u_a, u_b)$ from $Q$.
    \FOR {all arcs pairs $e_a$ and $e_b$ leaving $u_a$ and $u_b$} \label{alg:line:arc_loop}
        \STATE Get the output label $o_a$ of $e_a$.
        \STATE Get the input label $i_b$ of $e_b$.
        \IF {$o_a \ne i_b$}
            \STATE Continue to the next arc pair.
        \ENDIF
        \STATE Get destination states $v_a$ of $e_a$ and $v_b$ of $e_b$.
        \IF {$(v_a, v_b)$ is not in $R$}
            \STATE Continue to the next arc pair.
        \ENDIF
        \IF {$(v_a, v_b)$ is not in $\gC$}
            \STATE Add $(v_a, v_b)$ as a state to $\gC$ and to $Q$.
            \IF {$v_a$ and $v_b$ are accept states}
                \STATE Make $(v_a, v_b)$ an accept state in $\gC$.
            \ENDIF
        \ENDIF
        \STATE Get the weights $w_a$ of $e_a$ and $w_b$ of $e_b$.
        \STATE Add an arc to $\gC$ from $(u_a, u_b)$ to $(v_a, v_b)$ with label
        $o_a\!:\!i_b$ and weight $w_a\!+\!w_b$.
    \ENDFOR
\ENDWHILE
\STATE \textbf{Return}: The composed graph $\gC$.
\end{algorithmic}
\end{algorithm}

\subsection{Parallel Data Structure}
\label{sec:parallel_data}

We use a structure of arrays (SoA) layout for the transducer data structure.
The SoA layout is more efficient to use with SIMD-style algorithms than the
alternative array of structures, which is used for the sequential CPU
implementation.

\begin{figure}
    \centering
    \includegraphics[width=0.8\linewidth]{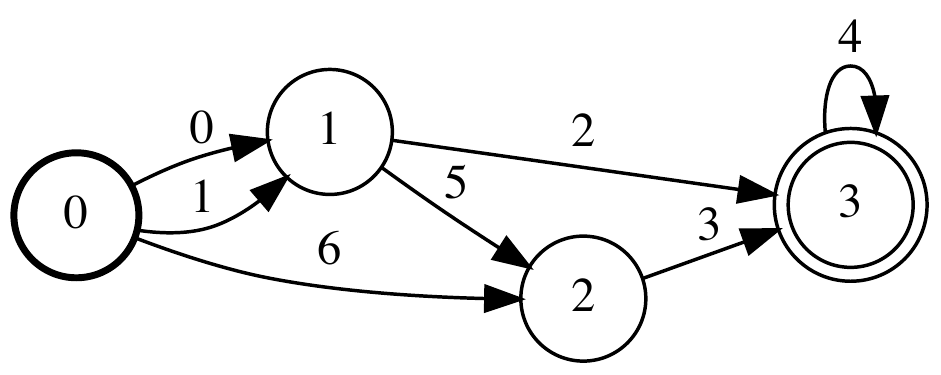}
    \caption{An example FST with edges labeled with their indices into the SoA
    representation.}
    \label{fig:example_fst}
\end{figure}

Consider an arbitrary transducer with $V$ nodes and $E$ edges. We store the
start and accept status of each node in two arrays with $V$ boolean entries. An
example for the graph in figure~\ref{fig:example_fst} is below:
\begin{verbatim}
  start   = {T, F, F, F}
  accept  = {F, F, F, T}
\end{verbatim}
Each node also contains a set of incoming edges and a set of outgoing edges.
We store the edge data in five arrays each with $E$ entries containing 1) the
input labels, 2) the output labels, 3) the weights, 4) the input node indices,
and 5) the output node indices. For each node, the indices of its input and
output edges in these five arrays are stored in two more arrays ({\tt inArcs}
and {\tt outArcs}). These two arrays are consecutive by node. For example, the
array of input arcs starts with the indices of incoming arcs to node $0$,
followed by the those for node $1$, and so on. The starting point for each
node in these arrays is stored in two more offset arrays ({\tt inArcOffset}
and {\tt outArcOffset}) with $V+1$ entries. The $v$-th entry states where the
$v$-th node's arcs begin, and the entry at $v+1$ is the index just after where
they end (or the start of the entries for node $v+1$).

\begin{figure*}
    \centering
    \begin{subfigure}{0.40\textwidth}
        \centering
        \includegraphics{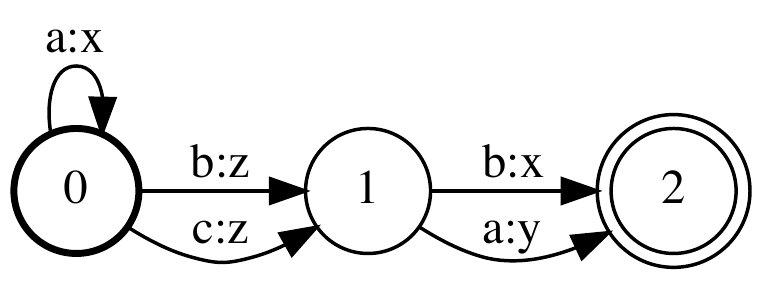}
    \end{subfigure}
    \hspace{10mm}
    \begin{subfigure}{0.40\textwidth}
        \centering
        \includegraphics{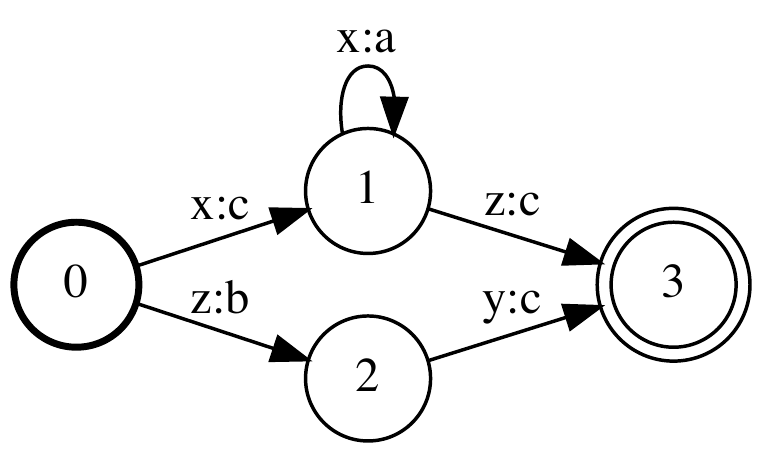}
    \end{subfigure}
    \caption{An example of the composition of two transducers. On the first
    step the queue contains the start state $Q = [(0, 0)]$, and $6$ arcs are
    explored.  On the second step, $Q = [(0, 1), (1, 2)]$, and $8$ arcs are
    explored. On the third step, $Q = [(1, 3), (2, 3)]$, and $0$ arcs are
    explored, after which the algorithm terminates. Notice that from the second
    to third step, the state $(0, 1)$ is reachable, but we do not add it to $Q$
    as it has already been considered.}
    \label{fig:compose}
\end{figure*}

An example of how incoming arcs are stored along with the input and output
labels for the graph in figure~\ref{fig:example_fst} is below:
\begin{verbatim}
  inArcOffset  = {0, 0, 2, 4, 7}
  inArcs       = {0, 1, 5, 6, 2, 3, 4}
  inLabels     = {a, b, c, d, e, f, g}
  outLabels    = {t, u, v, w, x, y, z}
\end{verbatim}
Suppose we want to find the input and output labels of the incoming arcs of
node $2$. We first find the span of the array {\tt inArcs} for node $2$ by
reading the third and fourth entries of {\tt inArcOffset}. The values of these
entries are $2$ and $4$. This means that the indices of the input arcs for the
second node start at index $2$ in {\tt inArcs} and end just before index $4$.
So we know the input arcs for node $2$ have indices $5$ and $6$ in the arrays
storing the input labels, the output labels, and the weights. The input labels
for these arcs are ${\tt f}$ and ${\tt g}$, and the output labels are ${\tt y}$
and ${\tt z}$.

\subsection{Parallel Algorithm}

At a high-level the primary distinction between the sequential and parallel compose
are the loops in lines~\ref{alg:line:state_loop} and \ref{alg:line:arc_loop} of
algorithm~\ref{alg:sequential_compose}. At each iteration of
algorithm~\ref{alg:sequential_compose}, the next state pair from the queue $Q$
is removed (line~\ref{alg:line:state_loop}). The next loop
(line~\ref{alg:line:arc_loop}) is over the cross product of outgoing arcs for
that state pair. The parallel algorithm simultaneously explores all arc pairs
for all state pairs currently in the queue. A separate thread is assigned to
each arc pair to be explored. The co-accessible subroutine
(line~\ref{alg:line:coaccessible}) is made parallel in the same way. An example
demonstrating the states and arcs explored at each iteration of the parallel
algorithm is given in figure~\ref{fig:compose}.

Many threads can attempt to modify any of the arrays described in
section~\ref{sec:parallel_data} while the composed graph is being constructed.
We modify the algorithm to avoid these potential race conditions. After
computing the coaccessible states, the parallel implementation performs two
passes over all possible accessible states of the composed graph. In the first
pass, we compute the number of new nodes in the composed graph along with the
number of input and output arcs for each. The offset into the arrays containing
edge data for each node is known at this point. On the second pass, the
algorithm fills in the correct values for the new arcs in the correct location.
In the first pass over accessible states the number of input and output arcs
for a node is incremented atomically. Similarly, in the second pass the arc
index being written for a given node is also incremented atomically. Access to
all other data structures is thread safe and does not require atomic operations
or other synchronization primitives.

%% file: experiments.tex
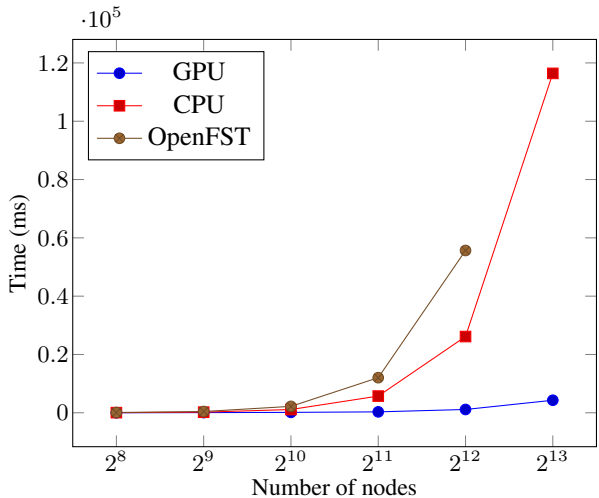
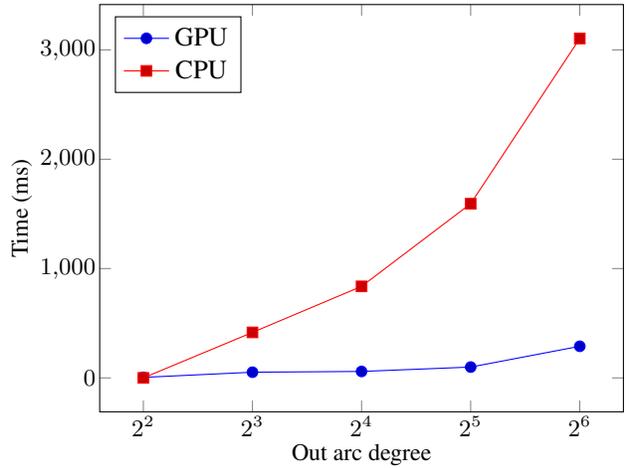
\begin{figure*}
    \centering
    \begin{subfigure}[b]{0.48\textwidth}
        \centering
        \input{figures/rand_nodes.tikz}
        \caption{Varying nodes.}
        \label{fig:rand_nodes}
    \end{subfigure}
    \hspace{4mm}
    \begin{subfigure}[b]{0.48\textwidth}
        \centering
        \input{figures/rand_arcs.tikz}
        \caption{Varying arc degree.}
        \label{fig:rand_arcs}
    \end{subfigure}
    \caption{A comparison of OpenFst CPU and GTN CPU and GPU
    composition on random graphs. (a) The number of nodes in the graph are
    increased while the outgoing arc degree and token set sizes are fixed at
    $5$ and $10$ respectively. (b) The outgoing arc degree is increased while
    keeping the number of nodes fixed at 256. In (b) the number of tokens is
    set to twice the outgoing arc degree to ensure the composition is not
    empty.}
    \label{fig:rand_bench}
\end{figure*}

\section{Experiments}
\label{sec:experiments}

In the following, we compare the GTN GPU and CPU implementations as well as an
OpenFst CPU implementation~\cite{allauzen2007openfst}. Code to reproduce these
benchmarks is open source and available at
\url{https://github.com/awni/parallel_compose}.  The CPU and GPU implementation
are open source as part of the GTN framework available at
\url{https://github.com/gtn-org/gtn}. The GPU benchmarks are performed on 32GB
Nvidia V100 GPUs. The CPU benchmarks are single-threaded and performed on
2.20GHz Intel Xeon E5-2698 CPUs.

\subsection{Random Graphs}
\label{sec:random_graphs}

The random graphs are constructed by first specifying the number of nodes. Each
graph has a single start and a single accept state. Outgoing edges from each
node are added such that every node in the graph has the same outgoing arc
degree. The destination node of each arc is randomly chosen from any of the
nodes in the graph including the source node; hence self-loops are allowed. The
arc label is randomly chosen from a predefined token set.

Figure~\ref{fig:rand_bench} compares the performance of the OpenFst CPU and the
GTN CPU and GPU compose implementations. In figure~\ref{fig:rand_nodes}, we
vary the number of nodes in the input graphs from 256 to 8,192 while
keeping the outgoing arc degree and token set size fixed at 5 and 10
respectively. In figure~\ref{fig:rand_arcs}, the outgoing arc degree of each
node is increased from 4 to 64 while keeping the number of nodes fixed.
Each point in the figures is the mean over several trials, the exact number
depending on the size of the graphs. Both the CPU and GPU implementations
ultimately scale quadratically in the number of nodes. However, the GPU
implementation is much faster at larger graph sizes. When the input graphs have
8,192 nodes, the GPU implementation is nearly 30 times faster than the GTN
CPU implementation.

\subsection{Composition with a Lexicon}

To emulate a more realistic computation, we compose an emissions graph with a
lexicon. The lexicon is derived from the LibriSpeech
corpus~\cite{panayotov2015librispeech} and contains 200,000 word-to-phoneme
mappings using 69 phonemes. We intersect this with an emissions graph designed
to emulate the output of an acoustic model. The emissions graph is linear with
251 nodes and 69 arcs between each node. The arcs between each node represent
the phoneme scores for the corresponding frame. We use 251 nodes as this
corresponds to 10 seconds of audio assuming a 10 millisecond step size for
acoustic features and a factor of four frame rate reduction in the network. We
compose the closure of a graph representing the lexicon with the linear
emissions graph. The closure of the lexicon adds $\epsilon$ transitions, so
this benchmark requires support for $\epsilon$ in the composition.

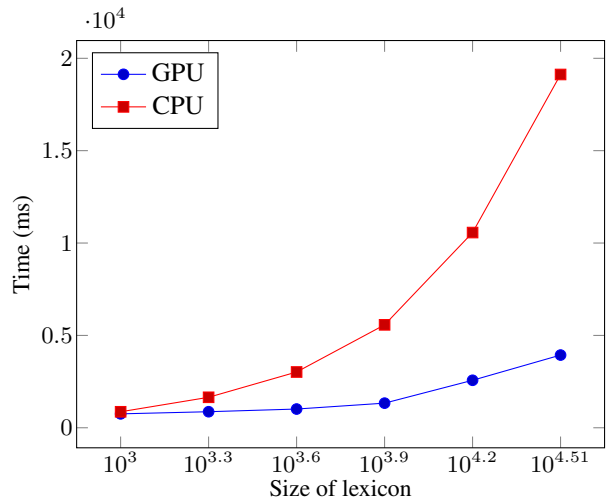
\begin{figure}
    \centering
    \input{figures/lexicon_compose.tikz}
    \caption{A comparison of the CPU and GPU composition with a lexicon and
    emissions graph for increasing number of words.}
    \label{fig:lexicon_compose}
    \vspace{-2cm}
\end{figure}

Figure~\ref{fig:lexicon_compose} compares the CPU and GPU composition while
increasing the number of words in the lexicon. The words are randomly sampled
without replacement. For a small number of words (1,000) the CPU and GPU
run times are comparable, but as we increase the number of words, the GPU
composition is much faster. At 32,000 words, the GPU implementation is more
than 10 times faster than the CPU implementation.

%% file: figures/rand_nodes.tikz
\begin{tikzpicture}[trim axis left, trim axis right]
\begin{axis}[
  height=7cm,
  inner sep=1.5,
  outer sep=0,
  xlabel={\small Number of nodes},
  ylabel={\small Time (ms)},
  xmode=log,
  log basis x={2},
  xtick=data,
  width=\linewidth,
  ylabel near ticks,
  xlabel near ticks,
  legend pos=north west,
  ticklabel style={font=\small},
]
\addplot table [y=gpu, x=num_nodes]{figures/rand_nodes_gtn.dat};
\addplot table [y=cpu_gtn, x=num_nodes]{figures/rand_nodes_gtn.dat};
\addplot table [y=cpu_openfst, x=num_nodes]{figures/rand_nodes_ofst.dat};
\legend{GPU, CPU, OpenFST};
\end{axis}
\end{tikzpicture}

%% file: figures/rand_arcs.tikz
\begin{tikzpicture}[trim axis left, trim axis right]
\begin{axis}[
  height=7cm,
  inner sep=1.5,
  outer sep=0,
  xlabel={\small Out arc degree},
  ylabel={\small Time (ms)},
  xmode=log,
  log basis x={2},
  xtick=data,
  width=\linewidth,
  ylabel near ticks,
  xlabel near ticks,
  legend pos=north west,
  ticklabel style={font=\small},
]
\addplot table [y=gpu, x=out_arc_degree]{figures/rand_arcs.dat};
\addplot table [y=cpu_gtn, x=out_arc_degree]{figures/rand_arcs.dat};
\legend{GPU, CPU};
\end{axis}
\end{tikzpicture}

%% file: figures/lexicon_compose.tikz
\begin{tikzpicture}[trim axis left, trim axis right]
\begin{axis}[
  height=7cm,
  inner sep=1.5,
  outer sep=0,
  xlabel={\small Size of lexicon},
  ylabel={\small Time (ms)},
  xmode=log,
  xtick=data,
  width=\linewidth,
  ylabel near ticks,
  xlabel near ticks,
  legend pos=north west,
  ticklabel style={font=\small},
]
\addplot table [y=gpu, x=words]{figures/lexicon_compose.dat};
\addplot table [y=cpu_gtn, x=words]{figures/lexicon_compose.dat};
\legend{GPU, CPU};
\end{axis}
\end{tikzpicture}

%% file: conclusion.tex
\section{Conclusion}
\label{sec:conclusion}

We presented an algorithm for parallel composition on GPUs. Our algorithm
handles general FSTs including $\epsilon$ transitions. For large graphs, the
parallel composition can be as much as $10$ to $30$ times faster than a highly
optimized sequential algorithm running on the CPU. We intend to continue to
refine and optimize the parallel composition. Developing parallel algorithms
which can handle $N$-way composition instead of just two inputs may yield
further improvements. Other important FST operations include determinize,
minimize, epsilon removal, and shortest path algorithms. A fully featured
framework for operations on FSTs which can run on GPUs will require parallel
implementations of these algorithms. Such a framework has the potential to open
the door for new and impactful machine-learning models built from FSTs.